\documentclass[letterpaper, 10 pt, conference]{ieeeconf}  

\IEEEoverridecommandlockouts                              

\usepackage{amsfonts}
\usepackage{amsmath}
\usepackage{bm}
\usepackage{graphicx}
\usepackage{bbm}
\usepackage{wrapfig}
\usepackage{booktabs, lipsum}
\usepackage{hyperref}
\usepackage{color}

\title{\LARGE \bf
Multimodal Safety-Critical Scenarios Generation for Decision-Making Algorithms Evaluation
}

\author{Wenhao Ding, Baiming Chen, Bo Li, Kim Ji Eun, Ding Zhao
\thanks{Wenhao Ding, Baiming Chen and Ding Zhao are with the Department of Mechanical Engineering, Carnegie Mellon University, USA (e-mail: \{wenhaod, baimingc, dingzhao\}@andrew.cmu.edu)}%
\thanks{Bo Li is with the Department of Computer Science, UIUC, USA (e-mail: lbo@illinois.edu)}%
\thanks{Kim Ji Eun is with Robert Bosch LLC (e-mail: JiEun.Kim@us.bosch.com)}%
}

\begin{document}

\maketitle
\thispagestyle{empty}
\pagestyle{empty}

\def\vzero{{\bm{0}}}
\def\vone{{\bm{1}}}
\def\vmu{{\bm{\mu}}}
\def\vtheta{{\bm{\theta}}}
\def\vphi{{\bm{\phi}}}
\def\vpsi{{\bm{\psi}}}
\def\va{{\bm{a}}}
\def\vb{{\bm{b}}}
\def\vc{{\bm{c}}}
\def\vd{{\bm{d}}}
\def\ve{{\bm{e}}}
\def\vf{{\bm{f}}}
\def\vg{{\bm{g}}}
\def\vh{{\bm{h}}}
\def\vi{{\bm{i}}}
\def\vj{{\bm{j}}}
\def\vk{{\bm{k}}}
\def\vl{{\bm{l}}}
\def\vm{{\bm{m}}}
\def\vn{{\bm{n}}}
\def\vo{{\bm{o}}}
\def\vp{{\bm{p}}}
\def\vq{{\bm{q}}}
\def\vr{{\bm{r}}}
\def\vs{{\bm{s}}}
\def\vt{{\bm{t}}}
\def\vu{{\bm{u}}}
\def\vv{{\bm{v}}}
\def\vw{{\bm{w}}}
\def\vx{{\bm{x}}}
\def\vy{{\bm{y}}}
\def\vz{{\bm{z}}}

\newcommand{\revisecolor}{black}

\begin{abstract}

Existing neural network-based autonomous systems are shown to be vulnerable against adversarial attacks, therefore sophisticated evaluation on their robustness is of great importance.
However, evaluating the robustness under the worst-case scenarios based on known attacks is not comprehensive, not to mention that some of them even rarely occur in the real world.
In addition, the distribution of safety-critical data is usually multimodal, while most traditional attacks and evaluation methods focus on a single modality. 
To solve the above challenges, we propose a flow-based multimodal safety-critical scenario generator for evaluating decision-making algorithms. 
The proposed generative model is optimized with weighted likelihood maximization and a gradient-based sampling procedure is integrated to improve the sampling efficiency.
\textcolor{\revisecolor}{
The safety-critical scenarios are generated by efficiently querying the task algorithms and a simulator.}
Experiments on a self-driving task demonstrate our advantages in terms of testing efficiency and multimodal modeling capability.
We evaluate six Reinforcement Learning algorithms with our generated traffic scenarios and provide empirical conclusions about their robustness.

\end{abstract}

\section{Introduction}

\textcolor{\revisecolor}{
Robustness and safety are crucial factors to determine whether a decision-making algorithm can be deployed in the real world~\cite{63}.} However, most of the data collected from simulations or in the wild are skewed to redundant and highly safe scenarios, which leads to the long tail problem~\cite{11}. Furthermore, a self-driving vehicle has to drive hundreds of millions of miles to collect safety-critical data~\cite{55}, resulting in expensive development and evaluation phases. Meanwhile, a large number of safe Reinforcement Learning (RL) algorithms~\cite{62} have been proposed recently, yet the evaluation of these algorithms mostly use uniform sampling scenarios, which have been proven to be insufficient due to poor coverage of rare risky events.

Adversarial attack~\cite{32, 33} is widely used to obtain specific examples when assessing the robustness of the model. This method only addresses extreme conditions, thus it does not provide comprehensive performance evaluations of the system. Researchers~\cite{36, 37} point out that there will always be loopholes in a neural network (NN) that can be attacked, hence, testing at different stress levels is deemed to provide more information about the robustness of the system. On the other hand, although the perturbation is limited during the attack, there is no guarantee that obtained samples are likely to occur in the real-world. It is a waste of resource to request robots to pass the tests that are unlikely to happen in practice.

\begin{figure}
\centering
\includegraphics[width=0.45\textwidth]{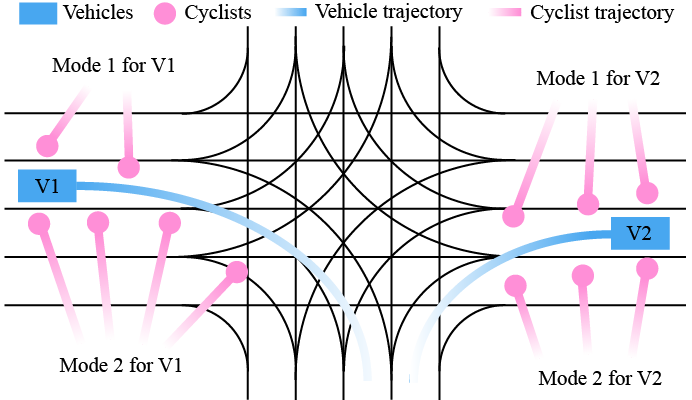}
\caption{Examples of multimodal safety-critical traffic scenarios. For each vehicle, there is a two-mode distribution of cyclist's spawn point that leads to high probability of collision. Pink trajectories represent samples from the distributions.
} 
\label{intro}
\end{figure}

The real-world scenarios are complicated with a huge number of parameters, and risk scenarios do not always happen within certain modality~\cite{39}. Multimodal distribution is a more realistic representation, for example, accidents could happen in different locations for a self-driving car as shown in Fig.~\ref{intro}. 
\textcolor{\revisecolor}{
Though previous works~\cite{17, 18} tried to search the risk scenarios under the RL framework, they only use the single-mode Gaussian distribution policy, leading to the over-fitting and unstable training problems. Covering diverse testing cases provides a more accurate comparison of algorithms. Even if a robot overfits to one specific risk modality, it will fail to handle other potential risk scenarios. To the best of our knowledge, few people have explored the multimodal estimation of safety-critical data.
}

In this paper, we use a flow-based generative model to estimate the multimodal distribution of safety-critical scenarios. We use the weighted maximum likelihood estimation (WMLE)~\cite{56} as the objective function, where the weight is related to the risk metric so that the log-likelihood of the sample will be approximately proportional to the risk level. We treat the algorithm that we want to evaluate as a black box, then get the risk value through the interaction with the simulation environment. To increase the generalization of generated scenarios, our generator also has a conditional input, so that the generated samples will be adaptively changed according to characteristics of the task.

We model the whole training process as an on-policy optimization framework which shares the same spirit as Cross Entropy Method (CEM)~\cite{70}, and we propose an adaptive sampler to improve the sample efficiency. Under this framework, we can dynamically adjust the region of interest of the sampler according to the feedback of the generator. The adaptive process is guided by the gradient estimated from the Natural Evolution Strategy (NES) method~\cite{30}. During the training stage, the sampler focuses on the unexplored and risky areas, and finally completes the multimodal modeling. We also consider the distribution of real-world data when designing the metric of risk to ensure that the generated data has a high probability of occurrence in the real world. 

We carried out extensive experiments on the decision-making task of autonomous driving in an intersection environment. A safety-critical generator is trained, analyzed, and compared with traditional methods. Our generator outperforms others in terms of the efficiency and multimodal covering capability. We also evaluate the robustness of several RL algorithms and claim that our generator is more informative than uniform sampling methods. In summary, the contribution is three-fold:
\begin{itemize}
	\item We propose a multimodal flow-based generative model that can generate adaptive safety-critical data to efficiently evaluate decision-making algorithms.
	\item We design an adaptive sampling method based on black-box gradient estimation to improve the sample efficiency of multimodal density estimation.
	\item We evaluate a variety of RL algorithms with our generated scenarios and provide empirical conclusions that can help the design and development of safe autonomous agents.
\end{itemize}


\section{Related Work}
\label{sec:related}

\subsection{Deep generative model}
Our method is based on deep generative models. The current popular generative models are mainly divided into three categories: normalizing flow~\cite{3} and autoregressive model~\cite{4} directly maximize the likelihood, Variational Auto-encoder (VAE)~\cite{2} optimizes the approximate likelihood using variational inference, and Generative adversarial network~\cite{1} implicitly computes the likelihood with a discriminator. The essence of these methods is to fit a distribution with parametric models that maximizes the likelihood according to the empirical data. In this paper, our data is collected from on-policy exploration. We select the flow-based model as the building block since we want to optimize the weighted likelihood directly and easily sample from the model.

\subsection{Adversarial Attack}
Another topic that is closely related to ours is the adversarial attack, which reduces the output accuracy of the target model by applying small disturbances to the original input samples. According to the information from the target model, this method falls into two types: white-box attack and black-box attack. Our method assumes that the internal information of the task algorithm cannot be obtained, so we are more relevant to the second one. This kind of method can be further divided into two mainstreams: substitute model~\cite{35} and query-based model~\cite{34}. The former is to train a completely accessible surrogate model to replace the target model, while the latter is based on the query of the target model to estimate the optimal attack direction. The NES gradient estimation method used in this paper has shown promising results in the latter method~\cite{38}.

\subsection{Safety-critical Scenario Generation}
Some previous works have used the generative model to conduct safety-critical scenarios search. \cite{16} modifies the last layer of a generative adversarial imitation learning model to generate different driving behaviors. \cite{14} and \cite{15} generate different levels of risky data by controlling the latent code of VAE. 
\textcolor{\revisecolor}{
Besides, some frameworks~\cite{17, 18, 21} combine RL and simulation environment to search for data that satisfies specific requirements. However, they only consider single-modal Gaussian policy.} Adaptive stress test~\cite{27, 28} is also a kind of methods using the Monte Carlo Tree Search and RL to generate collision scenarios. Most of these methods use the Gaussian distribution policy to describe the result of searching, without considering the case of multiple modality. 
In addition, lots of literatures borrow the idea from evolution algorithms \cite{67}, reinforcement learning \cite{69}, Bayesian optimization \cite{66}, and importance sampling \cite{64} to generate adversarial complex scenarios, resulting in diverse directions and platforms.
In previous works, \cite{19} is the most similar to this paper. They use the multilevel splitting method \cite{59} to gradually extract risk scenarios by squeezing the searching area. However, it is sensitive to level partition and the efficiency of Monte Carlo Markov Chain (MCMC) method is limited by query times and data dimension.

\subsection{Sampling Methods}
In online decision-making, especially RL tasks, exploration has always been a popular topic. The adaptive sampler proposed in this paper can also be categorized into this field. For the simple multi-arm bandit problem, traditional solutions are Upper Confidence Bound (UCB)~\cite{57} and Thompson sampling~\cite{58}. Recently, exploration methods based on curiosity~\cite{41}, and information gain~\cite{43} also cause much attention. Also, works like ~\cite{46} are based on the disagreement of ensemble models. To some extent, all these methods aim at modeling the environment and the explored area, so as to guide the sampler with the desired direction.

The gradient information usually facilities samplers faster convergence. Hamiltonian Monte Carlo (HMC) is a variant of MCMC method that is more efficient than vanilla random walk counterparts. Motivated by this, our proposed adaptive sampler also use a black-box estimator to obtain the gradient of a non-differential target function.


\begin{figure}
\centering
\includegraphics[width=0.46\textwidth]{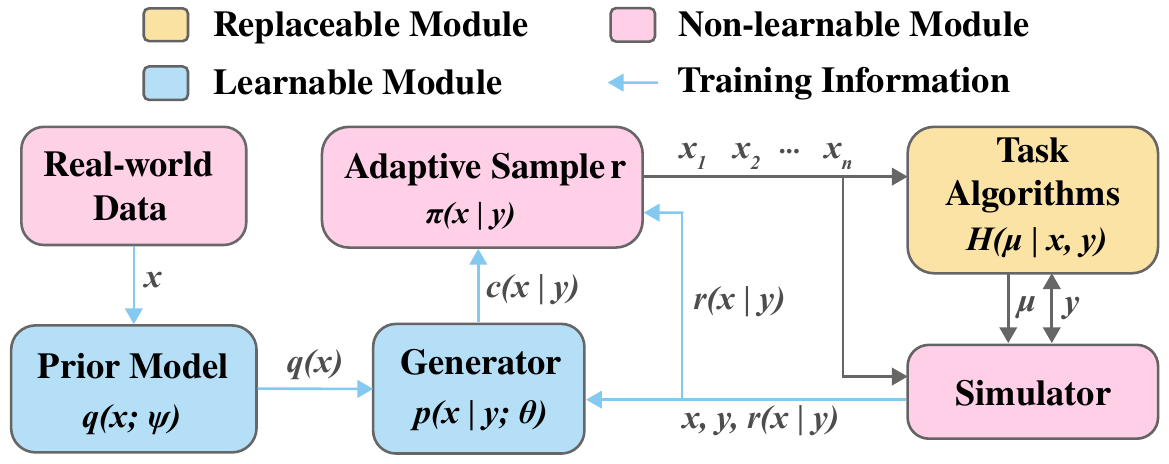}
\caption{Diagram of proposed framework.} 
\label{structure}
\end{figure}

\section{Proposed Method}
\label{sec:method}

\subsection{Notation and Preliminaries}

The variable $\vx \in \mathcal{X}$ represents parameters that build a scenario, for instance, the initial position of a pedestrian or the weather conditions in the self-driving context. The variable $\vy \in \mathcal{Y}$ represents the properties of the task, such as the goal position or target velocity. $H(\vmu|\vx, \vy)$ is the task algorithm that takes the scenario observation and task condition as input and outputs a decision policy $\vmu\in\mathcal{M}$. With a risk metric $r: \mathcal{X} \times \mathcal{M} \rightarrow \mathcal{R}$, each scenario is corresponding to a value that indicates the safety under $H$. For simplification, we omit the notation $\vmu$ in $r(\vx, \vmu|\vy)$.

Instead of exploring the extreme scenarios that output low $r(\vx|\vy)$ as in adversarial attack field, we aim at estimating a multimodal distribution $p(\vx|\vy)$ where the log-likelihood is proportional to the value of risk measure. Then we can efficiently generate scenarios that have different risk levels by sampling from $p(\vx|\vy)$. The condition $\vy$ follows a distribution $p(\vy)$. A sampler $\pi(\vx|\vy)$ is used to collect training data. We also assume real-world data of similar scenarios is accessible and has a distribution $q(\vx)$. The pipeline of our proposed evaluation method is shown in Fig.~\ref{structure}. We explain the details of three learnable modules in following sections.

\subsection{Pre-trained Prior Model}
\label{prior_model}

\textcolor{\revisecolor}{
Firstly, we consider the probability that each scenario happens in real-world to make the result practical. For that, we pre-train a generative model $q(\vx; \vpsi)$ to approximate the distribution of the real data $\mathcal{D}$. The objective of the training is maximizing the following log-likelihood:
\begin{equation}
	\hat{\vpsi} = \arg\max_{\vpsi} \mathbb{E}_{\vx\sim \mathcal{D}} \log q(\vx; \vpsi)
\end{equation}
}

\textcolor{\revisecolor}{
We select RealNVP~\cite{5}, a flow-based model to implement $q(\vx; \vpsi)$ for exact likelihood inference. In flow-based model, a simple distribution $p(\vz)$ is transformed to a complex distribution $p(\vx)$ by the change of variable theorem. Suppose we choose $\vz\sim \mathcal{N}(\vzero, \vone)$ to be the simple distribution. Then we have the following equation to calculate the exact likelihood of $\vx$:
\begin{equation}
	p(\vx) = p(\vz)\left| \frac{\partial \vz}{\partial \vx}\right| = p(f(\vx)) \left| \frac{\partial f(\vx)}{\partial \vx}\right|
\label{flow}
\end{equation}
where we have an invertible mapping $f: \mathcal{X} \rightarrow \mathcal{Z}$. For more details about the flow-based model, please refer to~\cite{5}. 
}

\textcolor{\revisecolor}{
After training, scenarios can be generated by $\vx = f^{-1}(\vz)$ where $z$ is sampled from $\mathcal{N}(\vzero, \sigma)$. A smaller $\sigma$ will make the samples more concentrated, therefore generated scenarios will have higher likelihood. Since our model is trained with WMLE, the high likelihood samples are also corresponding to high risk. Details about the network structure and hyperparameters can be found in Appendix.
}

\subsection{Generator}
\label{formulation}

\textcolor{\revisecolor}{
We formulate the safety-critical data generation as a density estimation problem. The traditional way to estimate the multimodal distribution $p(\vx|\vy)$ by a deep generative model is maximizing the likelihood of data. To integrate the risk information, we solve the problem by WMLE~\cite{56}. For one data point $\vx_i$, we have:
\begin{equation}
	L(\vx_i|\vy_{i}; \vtheta) = p(\vx_{i}|\vy_{i}; \vtheta)^{w(\vx_i)}
\end{equation}
\begin{equation}
	\log L(\vx_i|\vy_{i}; \vtheta) = {w(\vx_i|\vy)} \log p(\vx_{i}|\vy_{i}; \vtheta)
\end{equation}
where $w(\vx_i)$ is the weight and $p(\vx_i|\vy_i; \vtheta)$ is our generator with learnable parameter $\vtheta$, corresponding to the $i$-th data point. Assume we have a sampling distribution $\pi(\vx|\vy)$ of $\vx$, then our objective is:
\begin{equation}
	\hat{\vtheta} = \arg\max_{\vtheta} \mathbb{E}_{\vx\sim \pi(\vx|\vy), \vy\sim p(\vy)} \log L(\vx|\vy; \vtheta)
\label{objective}
\end{equation}
The definition of $w(\vx|\vy)$ is relevant to both $r(\vx|\vy)$ and $q(\vx; \vpsi)$:
\begin{equation}
	w(\vx_i) = r(\vx_i|\vy) + \beta q(\vx_i; \vpsi)
\end{equation}
where $\beta$ is a hyperparameter to balance $r(\vx)$ and $q(\vx; \vpsi)$. 
}

\textcolor{\revisecolor}{
We implement $p(\vx|\vy; \vtheta)$ by a modified flow-based model that has a conditional input~\cite{6}. Suppose $\vy \in \mathcal{Y}$ is the conditional input, then the mapping function should be $f: \mathcal{X} \times \mathcal{Y} \rightarrow \mathcal{Z}$ and (\ref{flow}) will be rewritten as:
\begin{equation}
	p(\vx|\vy) = p(f(\vx|\vy)) \left| \frac{\partial f(\vx|\vy)}{\partial \vx}\right|
\label{cflow}
\end{equation}
}

\subsection{Adaptive Sampler}
\label{sampler}

\textcolor{\revisecolor}{
The uniform distribution is a trivial choice for $\pi(\vx|\vy)$ in \ref{objective} to search the solution space. However, uniform sampling is inefficient in high-dimensional space, especially when the risky scenarios are rare. Therefore, we propose an adaptive sampler that leverages the gradient information to gradually cover all modes of risky scenarios. Suppose we have a metric $c(\vx|\vy)$ that indicates the exploration value: the higher $c(\vx|\vy)$ is, the more worth exploring $\vx$ is. We then use NES, a black-box optimization method, to estimate the gradient of the $c(\vx|\vy)$. The sampler $\pi(\vx|\vy)$ then follow the generating rule (we omit $\vy$ in gradient derivation):
\begin{equation}
	\vx^{t+1} \leftarrow \vx^{t} + \alpha \nabla_{\vx} c(\vx^{t})
\label{nes1}
\end{equation}
where $\alpha$ is the step size. The gradient $\nabla_{\vx} c(\vx^{t})$ in (\ref{nes1}) can be estimated by:
\begin{equation}
\begin{split}
	\nabla_{\vx} c(\vx^{t}) =& \nabla_{\vx} \mathbb{E}_{\vx \sim \mathcal{N}(\vx^{t}, \sigma^2 \bm{I})} \left[ c(\vx) \right]\\
							=& \frac{1}{\sigma} \mathbb{E}_{\epsilon \sim \mathcal{N}(\bm{0}, \bm{I})} \left[ \epsilon \cdot c(\vx^{t} + \sigma\epsilon) \right]
\end{split}
\label{nes}
\end{equation}
In practice, we will approximate the above expectation with the Monte Carlo method:
\begin{equation}
	\nabla_{\vx} c(\vx^{t}) = \frac{1}{\sigma} \sum_{i=1}^{M} \epsilon_{i} \cdot c(\vx^{t} + \sigma\epsilon_{i}),\ \ \ \ \epsilon_{i} \sim \mathcal{N}(\bm{0}, \bm{I})
\end{equation}
}

\begin{figure}
\centering
\includegraphics[width=0.47\textwidth]{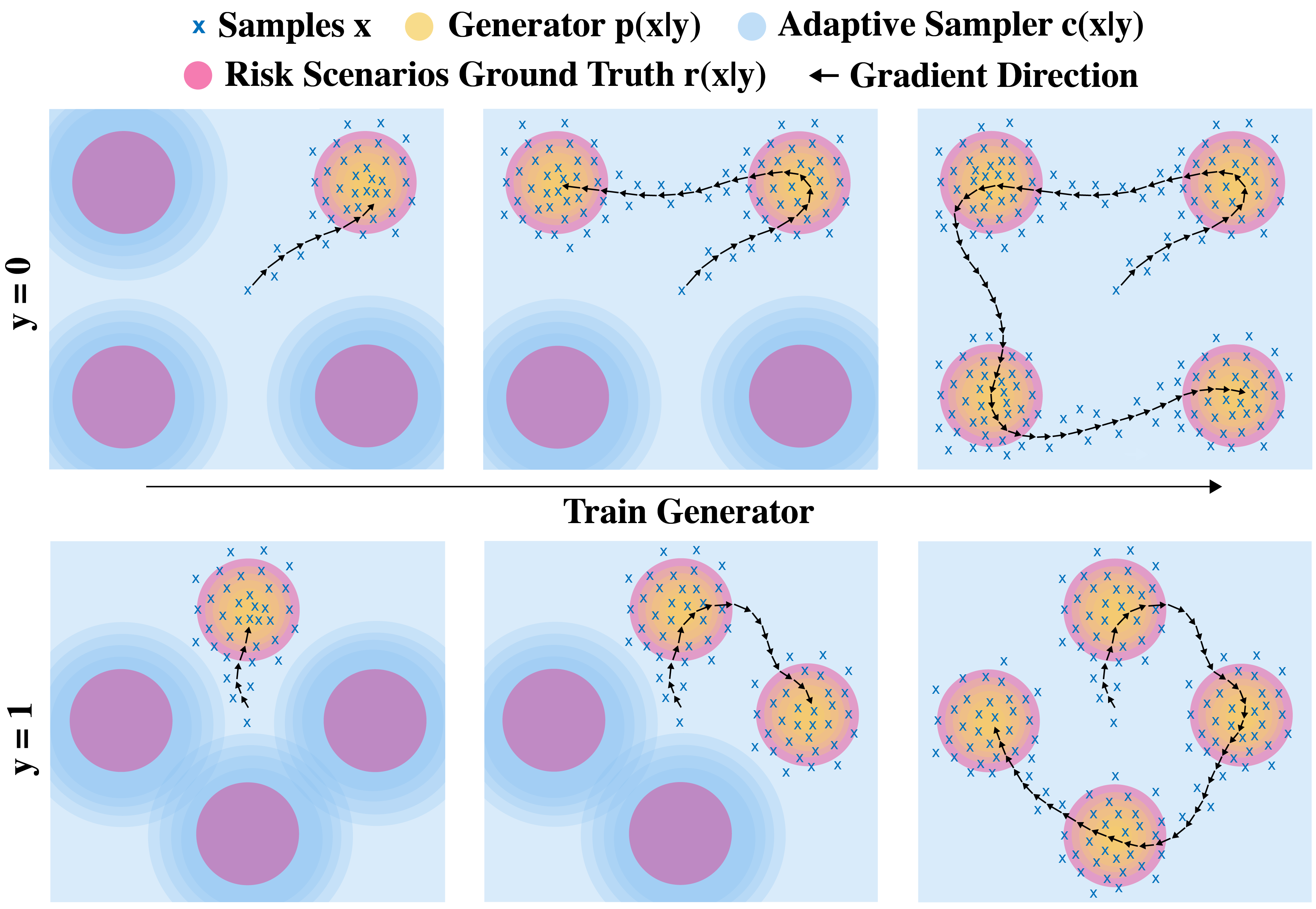}
\caption{
\textcolor{\revisecolor}{
A GMM example to illustrate our method. The condition $y$ has two values to simulate conditional distribution.The ground truth $r(x|y)$ is denoted by pink color, which is the likelihood of a 4-mode Gaussian distribution for each condition. The exploration value $c(x)$ is denoted by blue color, which gradually reduces as the samples expand. The generator $p(x)$ is denoted by yellow color, which gradually covers all modalities.
}}
\label{sampler_diagram}
\end{figure}

\textcolor{\revisecolor}{
The design of $c(\cdot)$ heavily influences the performance of the adaptive sampler. Inspired by some curiosity-driven literature~\cite{41}, where the uncertainty, Bayesian surprise, and prediction error are used to guide the exploration, we choose a metric that involves the generative model $p(\vx|\vy)$:
\begin{equation}
	c(\vx|\vy) = r(\vx|\vy) - \gamma\cdot p(\vx|\vy; \vtheta)
\label{curiosity}
\end{equation}
where $\gamma$ is a hyperparameter that balance $r(\vx)$ and $p(\vx|\vy; \vtheta)$. Intuitively, when one mode (some similar risky scenarios) is well learned by $p(\vx|\vy; \vtheta)$, the metric $c(\vx)$ will decrease and force the sampler to explore other modes. Finally, the multimodal distribution will be captured by $p(\vx|\vy; \vtheta)$. The diagram of this pipeline is shown in Fig.~\ref{sampler_diagram} with a Gaussian Mixture Model (GMM) example.
}

\begin{figure}
\centering
\includegraphics[width=0.48\textwidth]{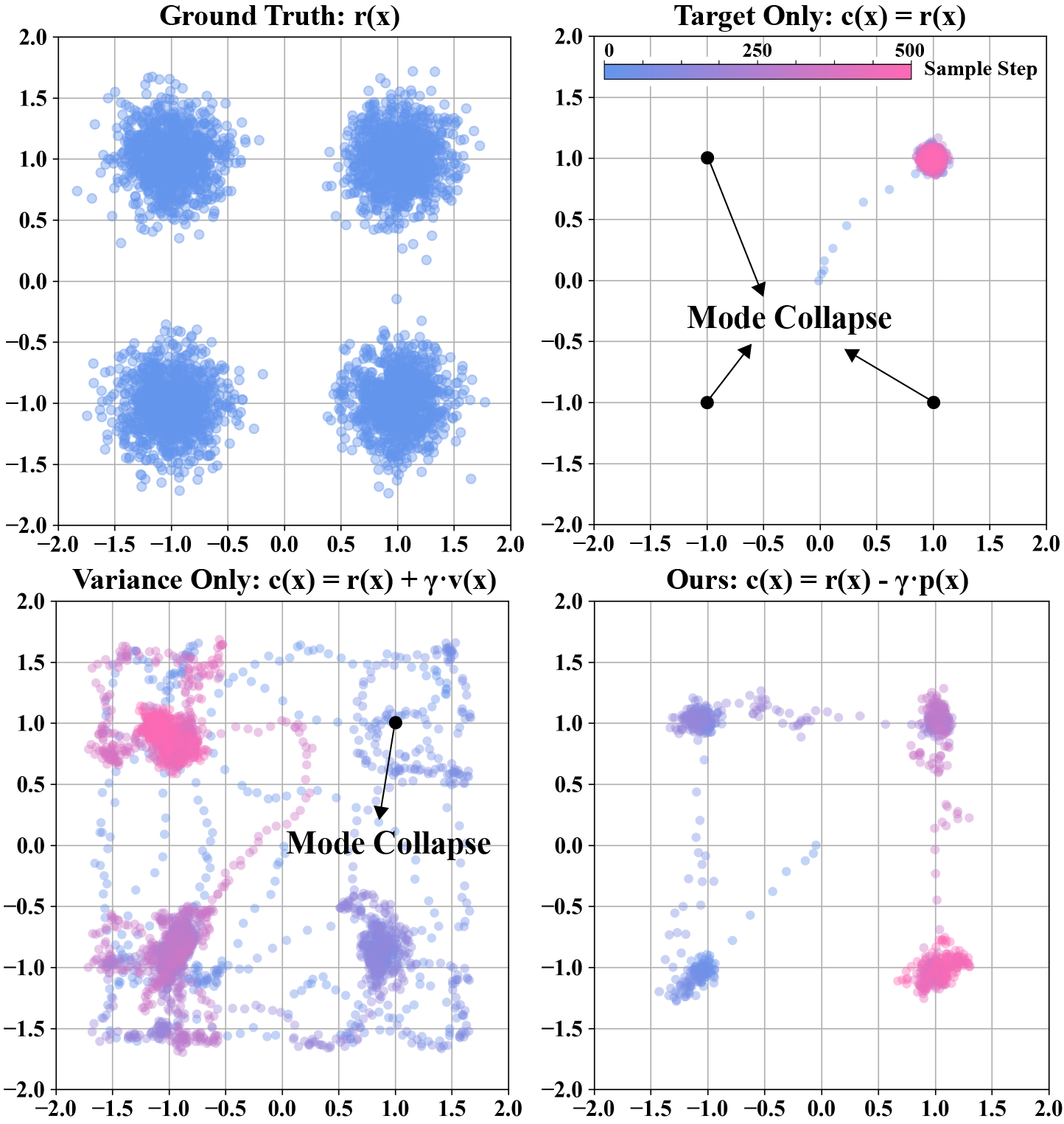}
\caption{A toy example to compare different adaptive samplers. The left-most figure shows the samples from $r(\vx)$, which is multimodal with four modalities. The other three figures show the samples obtained from three different samplers. Blue points indicate the early exploration stage and pink points indicate the later stage.} 
\label{gmm}
\end{figure}


\section{Experiments}
\label{sec:experiments}

In this section, we firstly demonstrate the advantage of our proposed adaptive sampler with a toy example. After that, we show the generated safety-critical scenarios with different settings in an intersection environment. Finally, we evaluate the robustness of several popular RL algorithms using our generated scenarios and provide conclusions about their robustness.

\subsection{Efficiency of Adaptive Sampler}

\textcolor{\revisecolor}{
\textbf{Environment settings.} As discussed in Section~\ref{formulation}, we shall expect that the estimated gradient of $c(\vx|\vy)$ improves the efficiency of the sampling procedure. To assess this, we compare our method with two baselines in a similar GMM example to Fig.~\ref{sampler_diagram} under $y=0$. In the first baseline, we use $c(\vx)=r(\vx)$, a straightforward and common choice in the adversarial attack literature. In the second baseline, we use the variance of the posterior of Gaussian Processes (GP) to model the uncertainty of search space and combine this uncertainty with $r(\vx)$. 
}

\textcolor{\revisecolor}{
\textbf{Explanation:} The comparison results is displayed in Fig.~\ref{gmm}. Both two baselines are facing the mode collapse problem to varying degrees, while our method effectively covers all modes. The reasoning is as follows. The first baseline uses only limited information about the multimodal landscape, thus is easily trapped into one modality. The second baseline, which gradually decreases the importance of the explored points, can cover all modes even other unimportant points. However, the rapidly descending uncertainty and the lack of adaptivity to the generator $p(\vx)$ lead to suboptimal results and unbalanced data collection. Our proposed method uses the feedback of the generator that gives the sampler both the capability of uncertainty exploration and balanced data collection, hence attaining all the modalities.
}

\subsection{Safety-critical Scenario Generation}

\textbf{Environment settings.} An intersection environment is used to conduct our experiment in the Carla simulator~\cite{50}. We represent the scenario as a 4-dimensional vector $\vx=[x, y, v_x, v_y]$, which represents the initial position and initial velocity of a cyclist. The cyclist is spawned in the environment and travels at a constant speed. Then we place an ego vehicle controlled by an intelligent driver model. \textcolor{\revisecolor}{The target route of the ego vehicle is represented by condition $y$.
The minimal distance between the cyclist and ego vehicle is used to calculate $r(\vx)$:
\begin{equation}
    r(\vx) = \text{exp}\{ -\| \vp_v - \vp_c \|_2 \}
\end{equation}
where $\vp_v$ and $\vp_c$ represents the position of the vehicle and the cyclist respectively.} 
A lower distance corresponds to a higher $r(\vx)$. This scenario is defined as a pre-crash scenario in~\cite{49} and also adopted in 2019 Carla Autonomous Driving Challenge~\cite{52}. This setting allows us to test the collision avoidance capability of decision-making algorithms $H$. Other more intelligent algorithms can replace the current agent during the evaluation stage.

\begin{figure}
\centering
\includegraphics[width=0.35\textwidth]{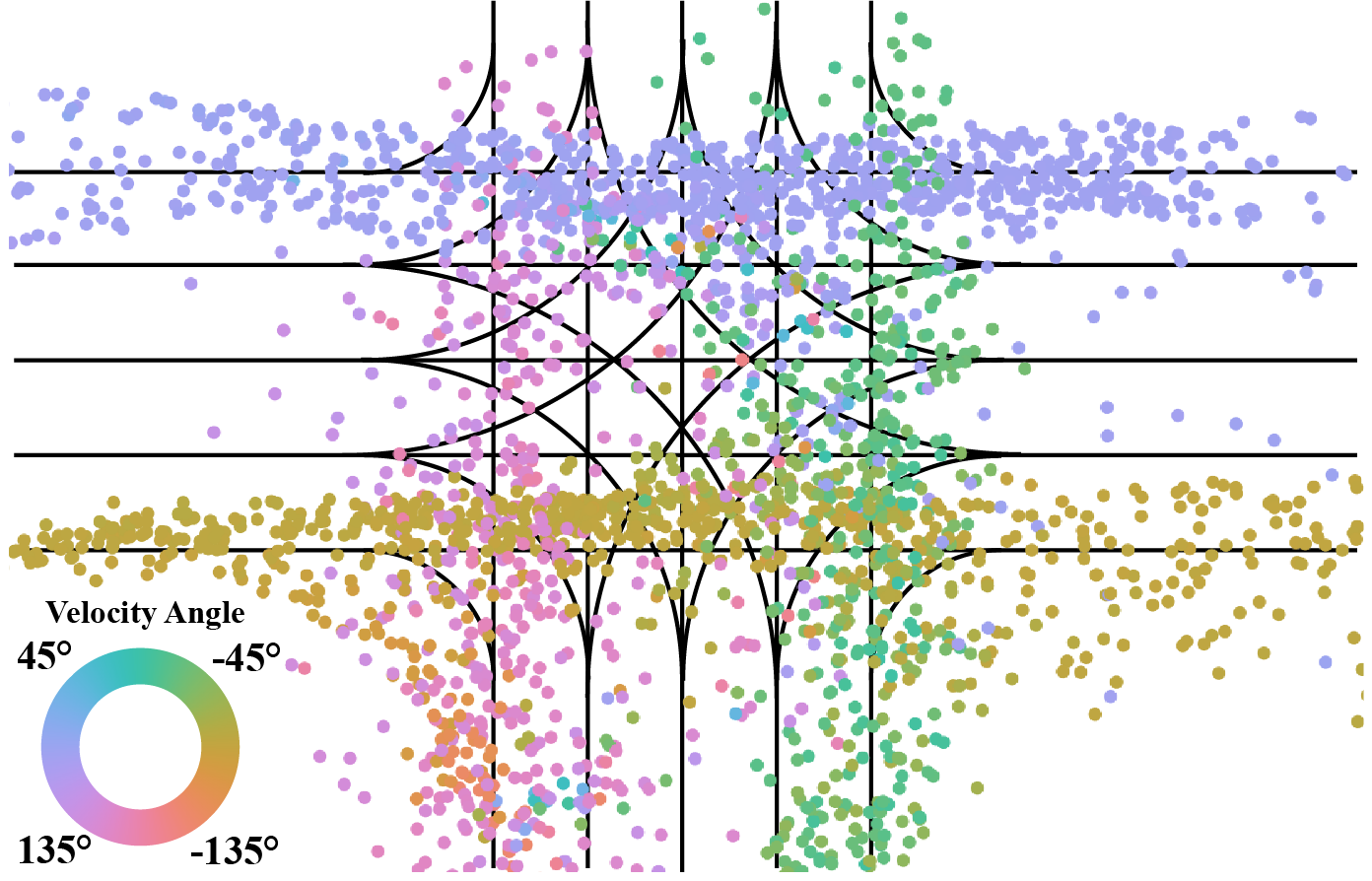}
\caption{Samples from prior model $q(\vx)$ that is learned from InD dataset~\cite{60}. Different colors represent different velocity angles.} 
\label{ind_dataset}
\end{figure}

\textbf{Real-world data distribution.} There are numerous datasets collected in the intersection traffic environment. We train our prior model $q(\vx)$ with trajectories from the InD dataset~\cite{60}. A well-trained prior model can be used to infer the likelihood of a given sample and generate new samples as well. We display the position and velocity direction of some generated samples in Fig.~\ref{ind_dataset}. These samples roughly describe the distribution of a cyclist in an intersection.

\begin{figure}
\centering
\includegraphics[width=0.48\textwidth]{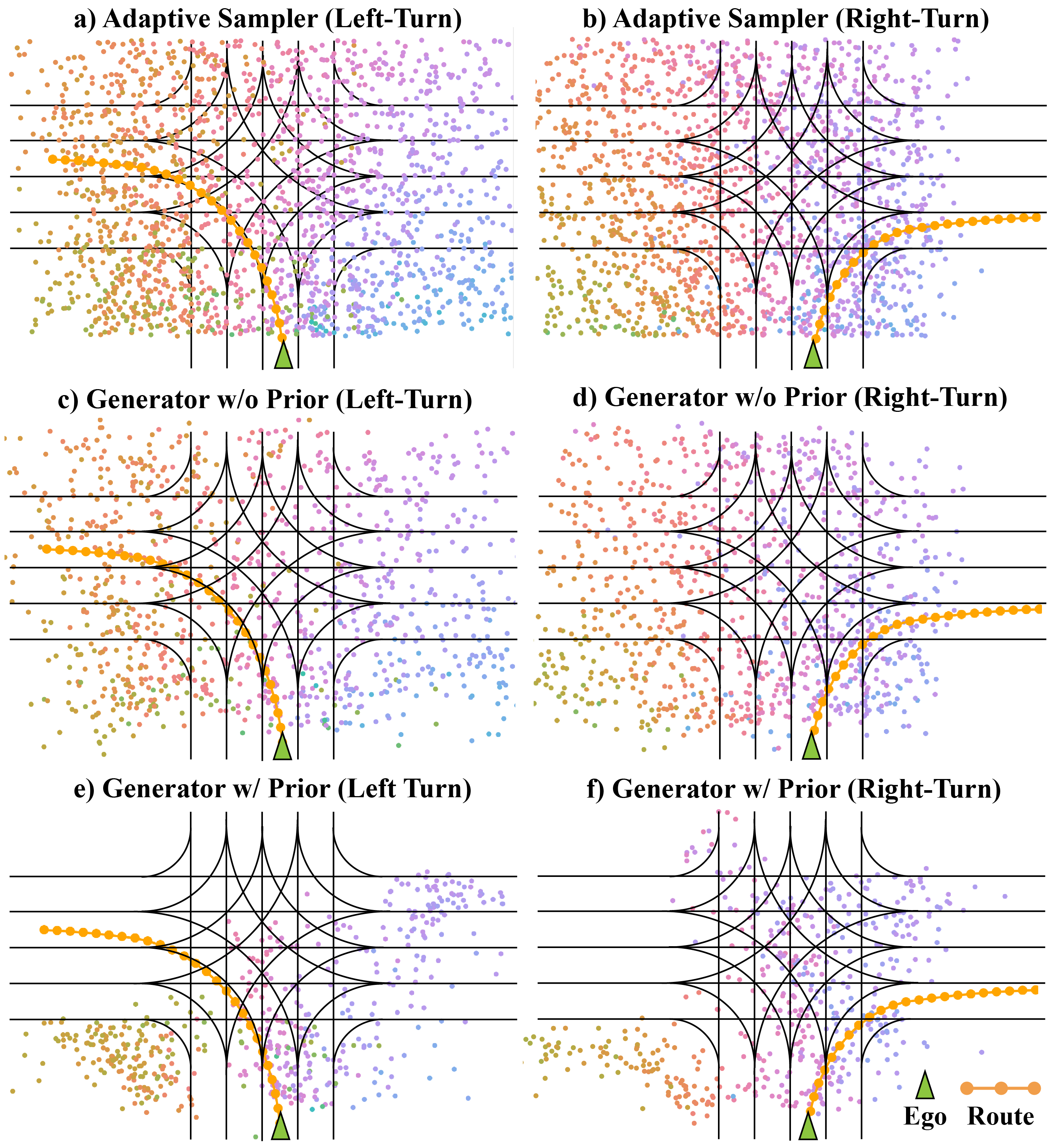}
\caption{Each point represents one risky scenario $\vx=[x, y, v_x, v_y]$. The color represents the direction of the velocity (same as Fig.~\ref{ind_dataset}). \textcolor{\revisecolor}{Condition $y$ represents the orange route.}
} 
\label{realnvp}
\end{figure}

\textcolor{\revisecolor}{\textbf{Generated scenarios display.} We train a generator $p(\vx|\vy)$ to generate safety-critical scenarios given the route condition $\vy$. In Fig.~\ref{realnvp}, we compare the samples from two generators: one does not use prior (middle row) and the other uses $q(\vx)$ as the prior model (bottom row), where the same color map is used as in Fig.~\ref{ind_dataset}. The top row of Fig.~\ref{realnvp} shows the collected scenarios by our adaptive sampler. Each of these samples is corresponding to a risk value that is not shown in the figure. In Fig.~\ref{realnvp}, it is shown that without real-world data prior, the generator learns the distribution of all risky scenarios collected by the adaptive sampler (first row). After incorporating the prior model (samples shown in Fig.~\ref{ind_dataset}), the generator concentrates more on the samples that are more likely to happen in the real world. This results in the removal of samples that has opposite directions to the real data.}

x

\textbf{Baseline and metric settings.} We select seven algorithms as our baseline. The details of each algorithm is discussed below:
\begin{itemize}
	\item Grid Search: We set the searching step for all parameters to 100. Since we $x$ has four dimensions, the entire searching iterations should be $10^8$.
	\item Human Design: We use the rules defined in \href{https://github.com/carla-simulator/scenario_runner}{Carla AD Challenge}, which basically trigger the movement of the cyclist according to the location of the ego vehicle.
	\item Uniform Sampling: The scenario parameter $x$ is uniformly sampled from the entire space. This method is widely used in the evaluation of safety decision-making algorithms. For instance, obstacles are randomly generated to test the collision avoidance performance in the Safety-Gym environment~\cite{61}.
	\item REINFORCE~\cite{18}: This method uses the REINFORCE framework with a single Gaussian distribution policy. This kind of policy can only represent single modality.
	\item REINFORCE+GMM: The policy distribution of~\cite{18} is replaced with a GMM. The purpose is to explore the multimodal capability of the REINFORCE algorithm.
	\item Ours-Uniform: We replace the adaptive sampler in our method with a uniform sampler.
	\item Ours-HMC: We replace the adaptive sampler in our method with a HMC sampler to explore the efficiency of gradient-based MCMC method. 
\end{itemize}
We use the query time and collision rate as our metrics. The query time means the number queries to the simulation during the training stage. Methods without training have 0 query time. A rough value is recorded when the distribution of samples is stable measured by human. The second metric is collision rate, which is calculated after the training stage. We sample 1000 scenarios for 10 different routes and get a collision rate for each route. We then calculate the mean and variance across the 10 routes.

\textbf{Comparison with baseline methods.} The results are shown in Table.~\ref{comparison}. Grid search is the most trivial way comparable with our method in finding the multimodal risk scenarios. However, the query time grows exponentially as the dimension of $\vx$ and the step size increase. 
In the simulation, human design is a possible way to reproduce the risk scenarios happen in the real world, but these scenarios are fixed and not adaptive to the changes of the task parameters $y$, leading to a low collision rate. 
Our experiment also found that the uniform method attains less than $10\%$ collision rate. In dense reward situations, uniform sampling could be a good choice, while in most real-world cases, the rare events risk scenarios makes this method quite inefficient. 
REINFORCE-Single searches the risk scenario under RL framework~\cite{18}. Although this method converges faster than ours, it cannot handle the multimodal cases with a single Gaussian distribution policy. The REINFORCE-GMM method extends the original version with a multimodal policy module. However, it has similar results as REINFORCE-Single. The reason is that on-policy sampling method in REINFORCE is easy to be trapped into a single modality, even the policy itself is multimodal. The final weight in GMM is highly imbalanced and only one component dominates.
The ablation study reveals that our adaptive sampler (Ours-Adaptive) is more efficient than the uniform version (Ours-Uniform). The MCMC version (Ours-HMC) requires less query time than our adaptive sampler, while its samples only concentrate on one modality.

\begin{figure}
\centering
\includegraphics[width=0.48\textwidth]{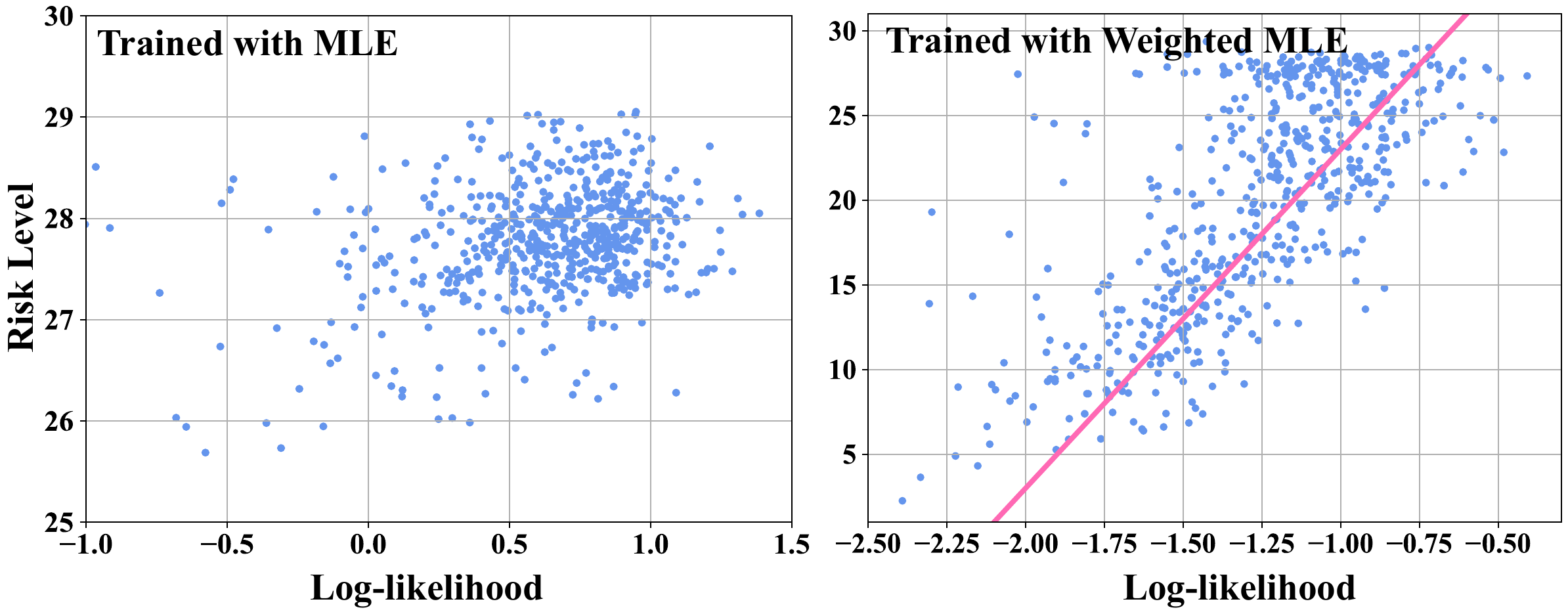}
\caption{Relationship between risk and log-likelihood of $p(\vx)$.} 
\label{linear}
\end{figure}

\textbf{Relationship between risk level and log-likelihood.} Since our generator is trained with WMLE, we make usage of all collected samples rather than only the risky ones as in~\cite{19}. We compare two generators that are trained with MLE and WMLE and plot the results in Fig.~\ref{linear}. The generator trained with MLE by only using the risk data concentrates on the high-risk area, while our WMLE generator has a linear relationship between the risk and log-likelihood. Therefore, our generator can not only generate risky scenarios but also generate scenarios with different risk levels by considering the likelihood of samples. 

\begin{figure*}
\centering
\includegraphics[width=0.96\textwidth]{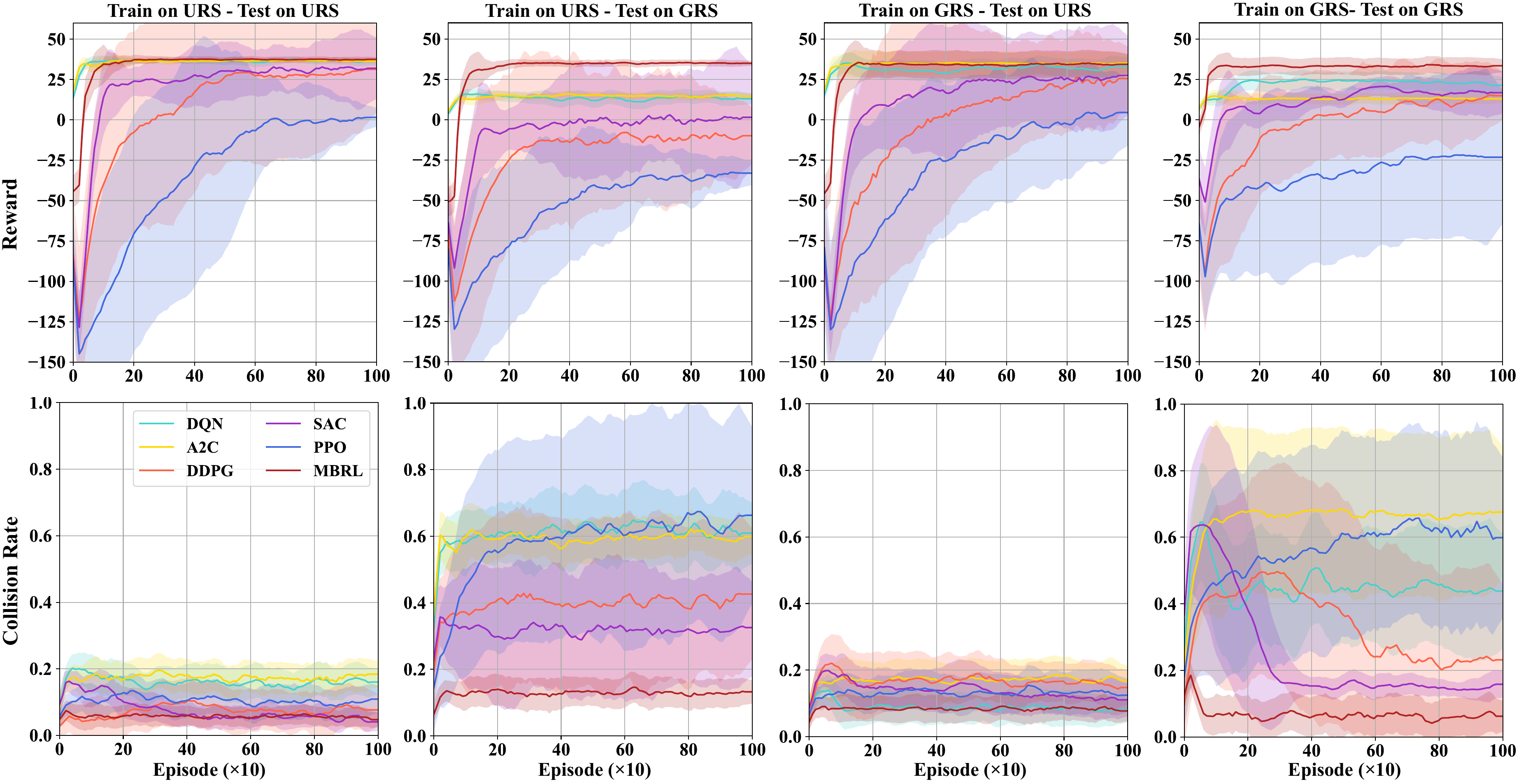}
\caption{Testing reward and testing collision rate in four different settings. Note that the action space of DQN and A2C is different from others, thus their results should not be compared to other methods. They are displayed together because they share the same reward space.} 
\label{rl}
\end{figure*}

\subsection{Evaluation of RL algorithms}

\textcolor{\revisecolor}{
To prove that our generated scenarios help improve the evaluation of algorithms, we implemented six popular RL agents (DQN~\cite{71}, A2C~\cite{72}, PPO~\cite{73}, DDPG~\cite{74}, SAC~\cite{75}, Model-based RL~\cite{76}) as $H(\mu|\vx, \vy)$ on the navigation task in the aforementioned environment. The target of agent is to arrive at a goal point $[x^g, y^g]$ and avoid reaching the non-driving area. At the same time, we place a cyclist on the intersection to create a traffic scenario. Finally, the state of the agent is:
\begin{equation}
    s = [x^g, y^g, x^a, y^a, v_x^a, v_y^a, x, y, v_x, v_y]
\end{equation}
where $[x^a, y^a, v_x^a, v_y^a]$ and $[x, y, v_x, v_y]$ represents the position and velocity of the agent and the cyclist, respectively.} The agents should also avoid colliding the cyclist otherwise they will receive a penalty. The reward consists of three parts:
\begin{equation}
	R(\vx) = r_{g} + r_{s}\times I_{s}(\vx) + r_{c}\times I_{c}(\vx)
\label{rk_reward}
\end{equation}
where $r_{g}$ is calculated by reduction of distance between the agent and the goal, $r_{s}$ and $r_{c}$ indicates the penalty of non-driving area violation and cyclist collision. $I_{s}(\vx)$ and $I_{c}(\vx)$ are two indicator functions that equal to 1 when the two events happen. The episode terminates when the agent collides into the cyclist or the agent reaches the target. We implement DQN and A2C on discrete action space with a controller that follows a pre-defined route. Their action space only influences the acceleration. The other agents have continuous action space that controls throttle and steering.  

\begin{table}
\centering
\caption{Performance Comparison}
\label{comparison}
\begin{tabular}{ccc}  
		\toprule 
		Methods              & Queries ($\downarrow$)   & Collision Rate ($\uparrow$) \\
		\midrule 
		Grid Search          & $1\times 10^8$         & $\bm{100\%}$\\
		Human Design         & -                      & $35\%\pm21\%$\\
		Uniform Sampling     & -                      & $9\%\pm1\%$\\
		REINFORCE-Single~\cite{18}  & $\bm{1\times 10^3}$    & $97\%\pm2\%$\\
		REINFORCE-GMM        & $\bm{1\times 10^3}$    & $98\%\pm1\%$\\
		
		Ours-Uniform                 & $1\times 10^5$         & $\bm{100\%}$\\
		Ours-HMC          & $\bm{1\times 10^3}$    & $\bm{100\%}$\\
		Ours-Adaptive                & $3\times 10^3$         & $\bm{100\%}$\\
		\bottomrule
\end{tabular}
\end{table}

\textcolor{\revisecolor}{
We have two environments for training and testing: 1) Uniform Risk Scenarios (URS): the initial state $\vx$ of the cyclist is uniformly sampled; 2) Generated Risk Scenarios (GRS): the initial state $\vx$ is sampled from our generated $p(\vx|\vy; \vtheta)$ with $\sigma=0.2$. We train and test six RL algorithms with different environments and Fig.~\ref{rl} displays the testing reward and testing collision rate. According to the comparison between different settings, we draw three main conclusions:
}

\textbf{Column 1 v.s. Column 2:} Agents tested on URS have similar final rewards and collision rates. These nearly indistinguishable results make it difficult to compare the robustness of different algorithms. In contrast, the results on GRS show a great discrepancy, which helps us obtain clearer conclusions. 

\textcolor{\revisecolor}{
\textbf{Column 1 v.s. Column 3:} We train the agents on URS and GRS but test them both on URS. We notice that the performance of both settings are similar, which means all agents do not sacrifice their generalization to URS.
}

\textcolor{\revisecolor}{
\textbf{Column 2 v.s. Column 4:} The expected results should be that all agents have improvement in column 4. However, we notice that different algorithms still show different robustness due to the their mechanisms. We roughly divide the six algorithms into three categories and explain them respectively. 
1) \textit{Improved a lot:} MBRL is robust to the risk scenario, even if it is not trained on GRS. The reason is that the target of MBRL is to learn a dynamics model and plan with it. Even if it is trained on normal scenarios, it learns how to predict the trajectory of the cyclist. Then when it is tested on risky scenarios, MBRL can easily avoid collision. Training on GRS will not make it perform better. 
2) \textit{Improved a little}: DDPG, DQN and SAC slightly improve the performance. From the bottom figure of column 4, we notice the collision rates of them firstly increase but quickly decrease, which means they learn how to deal with most of the risk scenarios. However, their rewards are lower than MBRL because they cannot handle all risky scenarios. The explanation for the little improvement is that these are off-policy methods with memory buffers. The average of stored risky scenarios from the buffer makes the training stable, therefore makes the agents successfully handle the scenarios they have met. However, they still fail in some unseen risky scenarios. 
3) \textit{Not Improved:} PPO and A2C are on-policy methods, which learn policy according to current samples. However, the risky scenarios cause the instability of training, because our generator finds different modes (types) of risky scenarios. In contrast, normal scenario will not cause such a problem because the state of the cyclist does not have much influence. In column 4, the collision rates of PPO and A2C gradually increase and never decrease, which means they cannot handle most risky scenarios.
}

Note that the above empirical conclusions might only be valid for in this environment. Further comparison of these RL algorithms should be carefully designed in multiple other settings. Nevertheless, our generator indeed is proven to be more insightful than the uniform sampler. Beyond RL algorithms, our proposed generating framework can also be used to efficiently evaluate other decision-making methods that are developed for dealing with more risky scenarios.


\section{Conclusion}
\label{sec:conclusion}

In this paper, we train a flow-based generative model using the objective function of weighted likelihood to realize the generation of multimodal safety-critical scenarios. Our generator can generate scenarios with various risk levels, providing efficient and diverse evaluations of decision-making algorithms. To speed up the training process, we propose an adaptive sampler based on feedback mechanism, which can adjust the sampling region according to the learning progress of the generator, and finally cover all risk modes in a faster way. We test six RL algorithms with scenarios generated by our generator in a navigation task and obtain some conclusions that are not easy to get with traditional uniform sampling evaluation. This achievement provides an efficient evaluation and comparison test-bed for the safety decision-making algorithms which have recently attracted more and more attention. A potential extension of this work is combining the evaluation and training process to build an adversarial training framework. We expect this combination can boost existing algorithms under safety-related tasks.


\section*{Acknowledgment}
The authors would like to thank Mansur Arief for helpful comments and discussion on drafts of this paper. This research was sponsored in part by Bosch.

\bibliographystyle{IEEEtran}
\bibliography{main}
\end{document}